\documentclass[10pt,twocolumn,letterpaper]{article}

\usepackage{cvpr}
\usepackage{times}
\usepackage{epsfig}
\usepackage{graphicx}
\usepackage{amsmath}
\usepackage{amssymb}
\usepackage{algorithm}
\usepackage{algorithmic}
\usepackage{subfigure}
\usepackage{amsmath}
\usepackage{mathtools}
\usepackage{multirow}
\usepackage{xcolor}
\usepackage{comment}
\usepackage{color}
\usepackage{etoolbox}
\newbool{inccomment}
\setbool{inccomment}{true}

\usepackage{amsmath,amsfonts,bm}

\def\eqref#1{equation~\ref{#1}}

\def\1{\bm{1}}

\def\vs{{\bm{s}}}

\def\evs{{s}}

\def\mI{{\bm{I}}}

\def\mK{{\bm{K}}}

\def\mU{{\bm{U}}}
\def\mV{{\bm{V}}}
\def\mW{{\bm{W}}}

\DeclareMathAlphabet{\mathsfit}{\encodingdefault}{\sfdefault}{m}{sl}
\SetMathAlphabet{\mathsfit}{bold}{\encodingdefault}{\sfdefault}{bx}{n}
\newcommand{\tens}[1]{\bm{\mathsfit{#1}}}

\def\tK{{\tens{K}}}

\def\sK{{\mathbb{K}}}

\newcommand{\R}{\mathbb{R}}

\newcommand{\normlzero}{L^0}
\newcommand{\normlone}{L^1}
\newcommand{\normltwo}{L^2}

\usepackage{url}
\usepackage{booktabs}
\usepackage{longtable}

\usepackage[pagebackref=true,breaklinks=true,letterpaper=true,colorlinks,bookmarks=false]{hyperref}

\cvprfinalcopy 

\def\cvprPaperID{4} 

\ifcvprfinal\fi
\begin{document}

\title{Learning Low-rank Deep Neural Networks via Singular Vector Orthogonality Regularization and Singular Value Sparsification}

\author{
Huanrui Yang\textsuperscript{\rm 1}, Minxue Tang\textsuperscript{\rm 2}, Wei Wen\textsuperscript{\rm 1}, Feng Yan\textsuperscript{\rm 3}, Daniel Hu\textsuperscript{\rm 4}, Ang Li\textsuperscript{\rm 1}, Hai Li\textsuperscript{\rm 1}, Yiran Chen\textsuperscript{\rm 1}\\ 
\textsuperscript{\rm 1}Department of Electrical and Computer Engineering, Duke University\\
\textsuperscript{\rm 2}Department of Electronic Engineering, Tsinghua University\\
\textsuperscript{\rm 3}Computer Science and Engineering Department, University of Nevada, Reno\\
\textsuperscript{\rm 4}Newport High School, Bellevue, WA\\
\textsuperscript{\rm 1}\{huanrui.yang, wei.wen, ang.li630, hai.li, yiran.chen\}@duke.edu, \textsuperscript{\rm 2}tangmx16@mails.tsinghua.edu.cn,\\ \textsuperscript{\rm 3}fyan@unr.edu, \textsuperscript{\rm 4}danielhu2003@gmail.com
}

\maketitle

\begin{abstract}
   Modern deep neural networks (DNNs) often require high memory consumption and large computational loads. 
In order to deploy DNN algorithms efficiently on edge or mobile devices, a series of DNN compression algorithms have been explored, including factorization methods.
Factorization methods approximate the weight matrix of a DNN layer with the multiplication of two or multiple low-rank matrices. 
However, it is hard to measure the ranks of DNN layers during the training process. 
Previous works mainly induce low-rank through implicit approximations or via costly singular value decomposition (SVD) process on every training step. 
The former approach usually induces a high accuracy loss while the latter has a low efficiency.
In this work, we propose \textbf{SVD training}, the first method to explicitly achieve low-rank DNNs during training without applying SVD on every step.
SVD training first decomposes each layer into the form of its full-rank SVD, then performs training directly on the decomposed weights. 
We add orthogonality regularization to the singular vectors, which ensure the valid form of SVD and avoid gradient vanishing/exploding. 
Low-rank is encouraged by applying sparsity-inducing regularizers on the singular values of each layer.
Singular value pruning is applied at the end to explicitly reach a low-rank model.
We empirically show that SVD training can significantly reduce the rank of DNN layers and achieve higher reduction on computation load under the same accuracy, comparing to not only previous factorization methods but also state-of-the-art filter pruning methods. 
\end{abstract}

\section{Introduction}
\label{sec:intro}

The booming development in deep learning models and applications has enabled beyond human performance in tasks like large-scale image classification~\cite{krizhevsky2012imagenet,he2016deep,hu2018squeeze,huang2017densely}, object detection~\cite{redmon2016you,liu2016ssd,he2017mask}, and semantic segmentation~\cite{long2015fully,chen2017deeplab}. 
Such high performance, however, comes with a high price of large memory consumption and computation load.
For example, a ResNet-50 model needs approximately $\mathrm{4G}$ floating-point operations (FLOPs) to classify a color image of $224 \times 224$ pixels. 
The computation load can easily expand to tens or even hundreds of GFLOPs for detection or segmentation models using state-of-the-art DNNs as backbones~\cite{canziani2016analysis}. 
This is a major challenge that prevents the deployment of modern DNN models on resource-constrained platforms, such as phones, smart sensors, and drones.

Model compression techniques for DNN models, including element-wise pruning~\cite{han2015learning,liu2015sparse,zhang2018systematic}, structural pruning~\cite{wen2016learning,luo2017thinet,li2019compressing}, quantization~\cite{liu2018bi,wang2019haq}, and factorization~\cite{jaderberg2014speeding,zhang2015accelerating,yang2015deep,xu2018trained}, have been extensively studied. 
Among these methods, quantization and element-wise pruning can effectively reduce model's memory consumption, but require specific hardware to realize efficient computation. 
Structural pruning reduces the computation load by removing redundant filters or channels. 
However, the complicated structures adopted in some modern DNNs (i.e., ResNet or DenseNet) enforce strict constraints on the input/output dimensions of certain layers. 
This requires additional filter grouping during the pruning and filter rearranging after the pruning to make the pruned structure valid~\cite{wen2017learning,ding2019centripetal}. 
Factorization method approximates the weight matrix of a layer with a multiplication of two or more low-rank matrices.
It by nature keeps the input/output dimension of a layer unchanged, and therefore the resulted decomposed network can be supported by any common DNN computation architectures, without additional grouping and post-processing.

The previous investigation show that it is feasible to approximate the weight matrices of a pretrained DNN model with the multiplication of low-rank matrices~\cite{jaderberg2014speeding,zhang2015accelerating,moczulski2015acdc,denton2014exploiting,lebedev2014speeding}. But these methods may greatly degrade the performance, even after post-hoc finetuning. 
Some other methods attempt to manipulate the ``directions'' of filters to implicitly reduce the rank of weight matrices~\cite{wen2017coordinating,li2019compressing}.
However, the difficulties in training and the implicitness of rank representation prevent these methods from reaching a high compression rate. 
Nuclear norm regularizer has been used to directly reduce the rank of weight matrices~\cite{alvarez2017compression,xu2018trained}.
Optimizing the nuclear norm requires conducting singular value decomposition (SVD) on every training step, which is inefficient, especially when dealing with larger models.

Our work aims to explicitly achieve a low-rank DNN network during the training without applying SVD on every step. 
In particular, we propose \textit{SVD training} by training the weight matrix of each layer in the form of its full-rank SVD. 
The weight matrix is decomposed into the matrices of left-singular vectors, singular values and right-singular vectors, and the training is done on the decomposed variables. 
Furthermore, two techniques are proposed to induce low-rank while maintaining high performance during the SVD training: 
(1) \textit{Singular vector orthogonality regularization} which keeps the singular vector matrices close to unitary throughout the training. 
It mitigates gradient vanishing/exploding during the training, and provide a valid form of SVD to guarantee the effective rank reduction.
(2) \textit{Singular value sparsification} which applies sparsity-inducing regularizers on the singular values during the training to induce low-rank.
The low-rank model is finally achieved through singular value pruning. 
We evaluate the individual contribution of each technique as well as the overall performance when putting them together via ablation studies. 
Results show that the proposed method constantly beats state of the art factorization and structural pruning methods on various tasks and model structures.
To the best of our knowledge, this is the first algorithm to explicitly search for the optimal rank of each DNN layer during the training without performing the decomposition operation at each training step.

\section{Related Works on low-rank DNNs}
\label{sec:relate}

Approximating a weight matrix with the multiplication of low-rank matrices is a straightforward idea for compressing DNNs. 
Early works in this field focus on designing the matrix and tensor decomposition scheme, especially for the 4-D tensor of convolution kernel, so that the operation of a pretrained network layer can be closely approximated with cascaded low-rank layers~\cite{jaderberg2014speeding,zhang2015accelerating,moczulski2015acdc,denton2014exploiting,lebedev2014speeding}. 
Tensor decomposition technique, notably CP-decomposition, is applied in early works to directly decompose the 4-D convolution kernel into 4 consecutive low-rank convolutions~\cite{lebedev2014speeding}. However, such decomposition technique significantly increases the number of layers in the achieved network, making them harder to be finetuned towards good performance, especially when decomposing larger and deeper models~\cite{tai2015convolutional}.
Later works therefore propose to reshape the 4-D tensor into a 2-D matrix, apply matrix decomposition technique like SVD to decompose the matrix, and finally reshape them back to 4-D tensors to get two consecutive layers. Notably, Zhang et al.~\cite{zhang2015accelerating} propose channel-wise decomposition, which uses SVD to decompose a convolution layer with a kernel size $w \times h$ into two consecutive layers with kernel sizes $w \times h$ and $1 \times 1$, respectively. 
The computation reduction can be achieved by exploiting the channel-wise redundancy, e.g., channels with smaller singular values in both decomposed layers are removed.
Similarly, Jaderberg et al.~\cite{jaderberg2014speeding} propose to decompose a convolution layer into two consecutive layers with less channels in between. 
They further utilize the spatial-wise redundancy to reduce the size of convolution kernels in the decomposed layers to $1 \times h$ and $w \times 1$, respectively.
These methods provide a closed-form decomposition for each layer. If the SVD is done in full rank, these methods guarantee that the decomposed layers perform the same operation as the original layer.
However, the weights of the pretrained model may not be low-rank by nature, so the manually imposed low-rank after decomposition by removing small singular values inevitably leads to high accuracy loss as the compression ratio increases~\cite{xu2018trained}.

Methods have been proposed to reduce the ranks of weight matrices during training process in order to achieve low-rank decomposition with low accuracy loss.
Wen et al.~\cite{wen2017coordinating} induce low rank by applying an ``attractive force'' regularizer to increase the correlation of different filters in a certain layer.
Ding et al.~\cite{ding2019centripetal} achieve a similar goal by optimizing with ``centripetal SGD,'' which moves multiple filters towards a set of clustering centers. 
Both methods can reduce the rank of the weight matrices without performing actual low-rank decomposition during the training.
However, the rank representations in these methods are implicit, so the regularization effects are weak and may lead to sharp performance decrease when seeking for a high speedup.
On the other hand, Alvarez et al.~\cite{alvarez2017compression} and Xu et al.~\cite{xu2018trained} explicitly estimate and reduce the rank throughout the training by adding Nuclear Norm (defined as the sum of all singular values) regularizer to the training objective. 
These method require performing SVD to compute and optimize the Nuclear Norm of each layer on every optimization step. 
Since the complexity of the SVD operation is $\mathcal{O}(n^3)$ and the gradient computation through SVD is not straightforward~\cite{giles2008extended}, performing SVD on every step is time consuming.

To explicitly achieve a low-rank network without performing costly decomposition on each training step, Tai et al.~\cite{tai2015convolutional} propose to directly train the network from scratch in the low-rank decomposed form, and add batch normalization~\cite{ioffe2015batch} between decomposed layers to tackle the potential gradient vanishing or exploding problem caused by the doubling of layers after decomposition.
However, the low-rank decomposed training scheme used in this line of works requires setting the rank of each layer before the training~\cite{tai2015convolutional}. 
The manually chosen low rank may not lead to the optimal compression. Also, training the low-rank model from scratch will make the optimization harder, as lower rank implies lower model capacity~\cite{xu2018trained}.

\section{Proposed Method}
\label{sec:method}

\begin{figure*}[tb]
    \centering
    \includegraphics[width=0.85\textwidth]{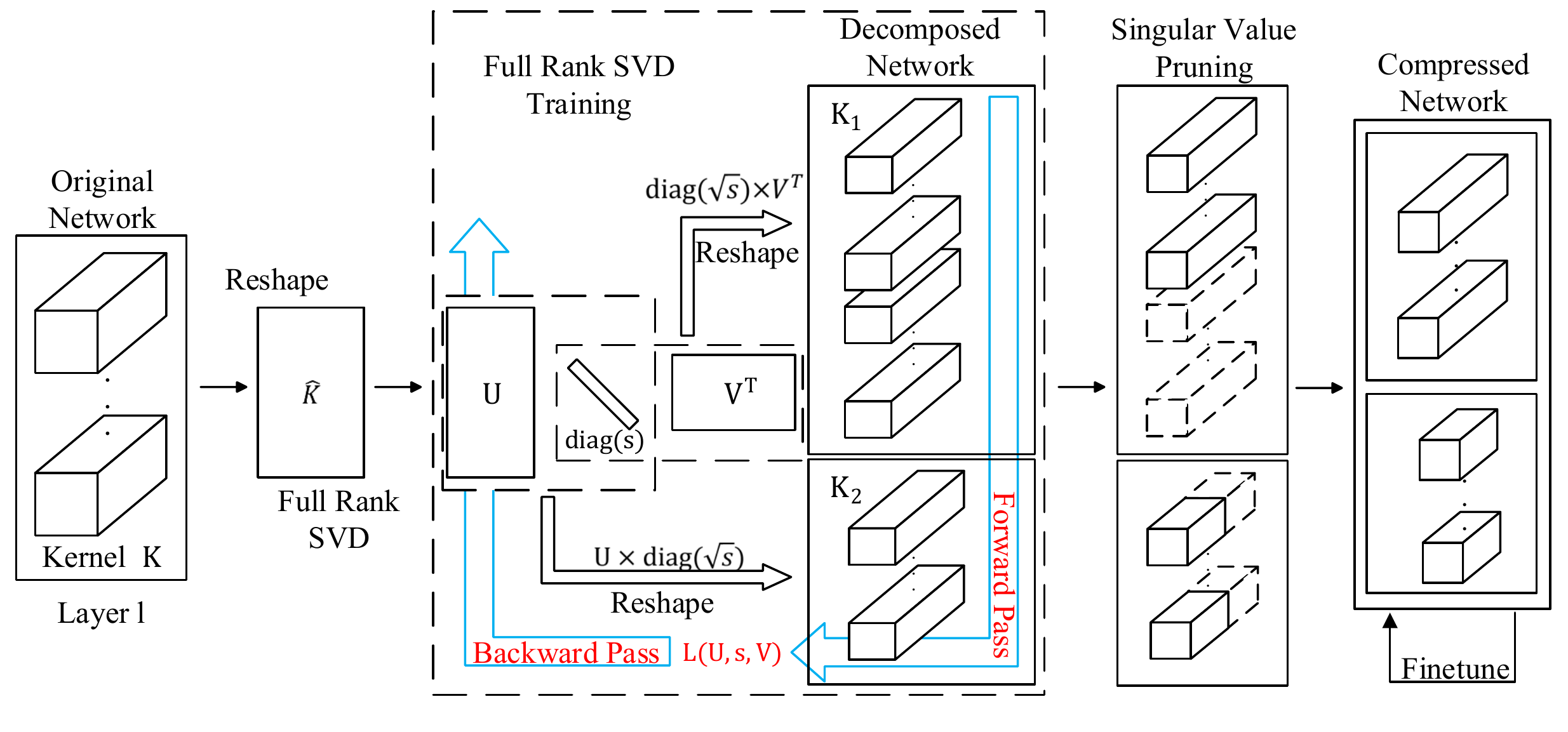}
    \vspace{-10pt}
    \caption{The training, compressing and finetuning pipeline of the proposed method.}
    \label{fig:pipeline}
    \vspace{-10pt}
\end{figure*}

Building upon previous works, we combine the ideas of decomposed training and trained low-rank in this work. 
As shown in Figure~\ref{fig:pipeline}, the model will first be trained in a decomposed form through the full-rank SVD training, then undergoes singular value pruning for rank reduction, and finally be finetuned for further accuracy recovery. 
As we will explain in Section~\ref{ssec:dec}, the model will be trained in the form of the spatial-wise~\cite{jaderberg2014speeding} or channel-wise decomposition~\cite{zhang2015accelerating} to avoid the time consuming SVD. 
Unlike the training procedure proposed by~\cite{tai2015convolutional}, we will train the decomposed model in its full-rank to preserve the model capacity. 
During the SVD training, we apply orthogonality regularization to the singular vector matrices and sparsity-inducing regularizers to the singular values of each layer, the details of which will be discussed in Section~\ref{ssec:orth} and~\ref{ssec:spar}, respectively. 
Section~\ref{ssec:train} will elaborate the full objective of the SVD training and the overall model compression pipeline.
This method is able to achieve optimal compression rate by inducing low-rank through training without the need for performing decomposition on every training step.

\subsection{SVD training of deep neural networks}
\label{ssec:dec}
In this work, we propose to train the neural network in its singular value decomposition form, where each layer is decomposed into two consecutive layers with no additional operations in between. 
For a fully connected layer, the weight $\mW$ is a 2-D matrix with dimension $\mW \in \R^{m\times n}$. 
Following the form of SVD, $\mW$ can be directly decomposed into three variables $\mU, \mV, \vs$ as $\mU \text{diag}(\vs) {\mV}^T$, with dimension $\mU \in \R^{m\times r}$, $\mV \in \R^{n\times r}$ and $\vs \in \R^{r}$. 
Both $\mU$ and $\mV$ shall be unitary matrices.
In the full-rank setting where $r = \min(m,n)$, $\mW$ can be exactly reconstructed as $\mW = \mU \text{diag}(\vs) {\mV}^T$. 
For a neural network, this is equivalent to decomposing a layer with weight $\mW$ into two consecutive layers with weight $\mW_1 = \mU \text{diag}(\sqrt{\vs})$ and $\mW_2 = \text{diag}(\sqrt{\vs}) {\mV}^T$ respectively.

For a convolution layer, the kernel $\tK$ can be represented as a 4-D tensor with dimension $\tK \in \R^{n\times c\times w \times h}$. 
Here $n, c, w, h$ represent the numbers of filters, the number of input channels, the width and the height of the filter respectively. 
This work mainly focuses on the \emph{channel-wise decomposition} method~\cite{zhang2015accelerating} and the \emph{spatial-wise decomposition} method~\cite{jaderberg2014speeding} to decompose the convolution layer, as these methods have shown their effectiveness in previous CNN decomposition research. 
For channel-wise decomposition, $\tK$ is first reshaped to a 2-D matrix $\hat{\mK}\in \R^{n\times cwh}$. 
$\hat{\mK}$ is then decomposed with SVD into $\mU \in \R^{n\times r}$, $\mV \in \R^{cwh\times r}$ and $\vs \in \R^{r}$, where $\mU$ and $\mV$ are unitary matrices and $r = \min(n,cwh)$. 
The original convolution layer is therefore decomposed into two consecutive layers with kernels $\tK_1 \in \R^{r\times c\times w \times h}$ reshaped from $\text{diag}(\sqrt{\vs}) {\mV}^T$ and $\tK_2 \in \R^{n\times r\times 1 \times 1}$ reshaped from $\mU \text{diag}(\sqrt{\vs})$.
Spatial-wise decomposition shares a similar process as the channel-wise decomposition. 
The major difference is that $\tK$ is now reshaped to $\hat{\mK}\in \R^{nw\times ch}$ and then decomposed into $\mU \in \R^{nw\times r}$, $\mV \in \R^{ch\times r}$, and $\vs \in \R^{r}$ with $r = \min(nw,ch)$. 
The resulting decomposed layers would have kernels $\tK_1 \in \R^{r\times c\times 1 \times h}$ and $\tK_2 \in \R^{n\times r\times w \times 1}$ respectively.
\cite{zhang2015accelerating} and \cite{jaderberg2014speeding} theoretically show that the decomposed layers can exactly replicate the function of the original convolution layer in the full-rank setting. 
Therefore training the decomposed model at full-rank should achieve a similar accuracy as training the original model.

During the SVD training, for each layer we use the variables from the decomposition, i.e., $\mU, \vs, \mV$, instead of the original kernel $\tK$ or weight $\mW$ as the trainable variables in the network. The forward pass will be executed by converting the $\mU, \vs, \mV$ into a form of the two consecutive layers as demonstrated above, and the back propagation and optimization will be done directly with respect to the $\mU, \vs, \mV$ of each layer. 
In this way, we can access the singular value $\vs$ directly without performing the time-consuming SVD on each step. 

Note that $\mU$ and $\mV$ need to be orthogonal to guarantee the low rank approximation can be done by removing small singular values, but this is not naturally induced by the decomposed training process. Therefore we add orthogonality regularization to $\mU$ and $\mV$ to tackle this problem as discussed in Section~\ref{ssec:orth}. Rank reduction is induced by adding sparsity-inducing regularizers to the $\vs$ of each layer, which will be discussed in Section~\ref{ssec:spar}.

\subsection{Singular vectors orthogonality regularizer}
\label{ssec:orth}
In a standard SVD procedure, the resulted $\mU$ and $\mV$ should be orthogonal by construction, which provides theoretical guarantee for the low-rank approximation. 
However, $\mU$ and $\mV$ in each layer are treated as free trainable variables in the decomposed training process, so the orthogonality may not hold. 
Without the orthogonal property, it is unsafe to prune the singular value in $\vs$ even if it reaches a small value, because the corresponding singular vectors in $\mU$ and $\mV$ may have high energy and lead to a large difference to the result. 

To make the form of SVD valid and enable effective rank reduction via singular value pruning, we introduce an orthogonality regularization loss to $\mU$ and $\mV$ as:
\begin{equation}
\label{equ:orth}
    L_o(\mU,\mV) = \frac{1}{r^2}(||{\mU}^T \mU-\mI||_F^2 + ||{\mV}^T \mV-\mI||_F^2),
\end{equation}  
where $||\cdot||_F$ is the Frobenius norm of matrix and $r$ is the rank of $\mU$ and $\mV$. 
Note that the ranks of $\mU$ and $\mV$ are the same given their definition in the decomposed training procedure.
Adding the orthogonality loss in Equation~(\ref{equ:orth}) to the total loss function forces $\mU$s and $\mV$s of all the layers close to be orthogonal matrices.

Beyond maintaining valid SVD form, the orthogonality regularization also bring additional benefit to the performance of the decomposed training process. The decomposed training process convert one layer in the original network to two consecutive layers, therefore doubles the number of layers. 
As aforementioned in~\cite{tai2015convolutional}, this may worsen the problem of exploding or vanishing gradient during the optimization, degrading the performance of the achieved model. 
Since the proposed orthogonality loss can keep all the columns of $\mU$ and $\mV$ to have the $\normltwo$ norms close to $1$, it can effectively prevent the gradient to explode or vanish when passing through variable $\mU$ and $\mV$, therefore helping the training process to achieve a higher accuracy. 
The accuracy gain brought by training with the orthogonality loss will be discussed in our ablation study in Section~\ref{ssec:exporth}.

\subsection{Singular values sparsity-inducing regularizer}
\label{ssec:spar}

With orthogonal singular vector matrices, reducing the rank of the decomposed network is equivalent to making the singular value vector $\vs$ of each layer sparse.
Although the sparsity of a vector is directly represented by its $\normlzero$ norm, it is hard to optimize the norm through gradient-based methods. 
Inspired by the recent works in DNN pruning~\cite{liu2015sparse,wen2016learning}, we use differentiable sparsity-inducing regularizer to make more elements in $\vs$ closer to zero, and apply post-train pruning to make the singular value vector sparse.

For the choice of the sparsity-inducing regularizer, the $\normlone$ norm has been commonly applied in feature selection~\cite{tibshirani1996regression} and DNN pruning~\cite{wen2016learning}. The $\normlone$ regularizer takes the form of $\normlone(\vs) = \sum_i |\evs_i|$, which is both almost everywhere differentiable and convex, making it friendly for optimization. 
Moreover, applying $\normlone$ regularizer on the singular value $\vs$ is equivalent to regularizing with the nuclear norm of the original weight matrix, which is a popular approximation to the rank of a matrix~\cite{xu2018trained}. 

However, the $\normlone$ norm is proportional to the scaling of parameters, i.e., $||\alpha W||_1 = \alpha ||W||_1$, with a non-negative constant $\alpha$. 
Therefore, minimizing the $\normlone$ norm of $\vs$ will shrink all the singular values simultaneously. 
In such a situation, some singular values that are close to zero after training may still contain a large portion of the matrix's energy.
Pruning such singular values may undermine the performance of the neural network.

To mitigate the proportional scaling problem of the $\normlone$ regularizer, previous works in compressed sensing have been using Hoyer regularizer to induce sparsity in solving non-negative matrix factorization~\cite{hoyer2004non} and blind deconvolution~\cite{krishnan2011blind}, where the Hoyer regularizer shows superior performance comparing to other methods. 
The Hoyer regularizer is formulated as 
\begin{equation}
    \label{equ:Hoyer}
    L^H(\vs) = \frac{|| \vs ||_1}{|| \vs ||_2}=\frac{\sum_i |\evs_i|}{\sqrt{\sum_i \evs_i^2}},
\end{equation}
which is the ratio of the $\normlone$ norm and the $\normltwo$ norm of a vector~\cite{krishnan2011blind}.
It can be easily seen that the Hoyer regularizer is almost everywhere differentiable and scale-invariant. 
The differentiable property implies that the Hoyer regularizer can be easily optimized as part of the objective function.
The scale-invariant property shows that if we apply the Hoyer regularizer to $\vs$, the total energy will be retained as the singular values getting sparser. Therefore most of the energy will be kept within the top singular values while the rest getting close to zero. 
This makes Hoyer regularizer attractive in our training process. 
The effectiveness of the $\normlone$ regularizer and the $L^H$ regularizer is explored and compared in Section~\ref{ssec:expspar}.

\subsection{Overall objective and training procedure}
\label{ssec:train}

With the analysis above, we propose the overall objective function of the decomposed training as:
\begin{equation}
    \begin{split}
    \label{equ:obj}
    L(\mU, \vs, \mV) = &L_T(\text{diag}(\sqrt{|\vs|}) {\mV}^T, \mU \text{diag}(\sqrt{|\vs|})) \\
    &+ \lambda_o \sum_{l=1}^D L_o(\mU_l, \mV_l) + \lambda_s \sum_{l=1}^D L_s(\vs_l).
\end{split}
\end{equation}

Here $L_T$ is the training loss computed on the model with decomposed layers. 
$L_o$ denotes the orthogonality loss provided in Equation~(\ref{equ:orth}), which is calculated on the singular vector matrices $\mU_l$ and $\mV_l$ of layer $l$ and added up over all $D$ layers. 
$L_s$ is the sparsity-inducing regularization loss, applying to the vector of singular values $\vs_l$ of each layer. 
We explore the use of both the $\normlone$ regularizer and the $L^H$ regularizer (Equation~(\ref{equ:Hoyer})) as $L_s$ in this work.
$\lambda_s$ and $\lambda_o$ are the decay parameters for the sparsity-inducing regularization loss and the orthogonality loss respectively, which are hyperparameters of the proposed training process. 
$\lambda_o$ can be chosen as a large positive number to enforce the orthogonality of singular vectors, and $\lambda_s$ can be modified to explore the tradeoff between accuracy and FLOPs of the achieved low-rank model.

As shown in Figure~\ref{fig:pipeline}, the low-rank decomposed network will be achieved through a three-stage process of \emph{full-rank SVD training}, \emph{singular value pruning} and \emph{low-rank finetuning}. 
First we train a full-rank decomposed network using the objective function in Equation~(\ref{equ:obj}). 
Training at full rank enables the decomposed model to easily reach the performance of the original model, as there is no capacity loss during the full-rank decomposition.
With the help of the sparsity-inducing regularizer, most of the singular values will be close to zero after the full-rank training process. 
Inspired by~\cite{xu2018trained}'s work, we prune the singular values using an energy based threshold. 
For each layer we find a set $\sK$ with the largest number of singular values subject to:
\begin{equation}
    \sum_{j\in \sK} \evs_j^2 \leq e \sum_{i=1}^r \evs_i^2,
\end{equation}
where $e\in[0,1]$ is a predefined energy threshold. We use the same threshold for all the layers in our experiments.
When $e$ is small enough, the singular values in set $\sK$ and the corresponding singular vectors can be removed safely with negligible performance loss.
The pruning step will dramatically reduce the rank of the decomposed layers. 
For a convolution layer with kernel $\tK \in \R^{n\times c\times w \times h}$, if we can reduce the rank of the decomposed layers to $r$, the number of FLOPs for the convolution will be reduced by $\frac{(n+chw)r}{nchw}$ or $\frac{(nw+ch)r}{nchw}$ when channel-wise or spatial-wise decomposition is applied, respectively.
The resulted low-rank model will then be finetuned with $\lambda_s$ set to zero for further performance recovery.

\section{Experiment Results}
\label{sec:exp}

In this section, we first perform ablation studies on the effectiveness of each part of our training procedure using ResNet models~\cite{he2016deep} on the CIFAR-10 dataset~\cite{krizhevsky2009learning}. 
We then apply the proposed decomposed training method on various DNN models on the CIFAR-10 dataset and the ImageNet ILSVRC-2012 dataset~\cite{ILSVRC15}. 
The training hyperparameters for these models can be found in Appendix~\ref{ap:setup}.
Different hyperparameters are used to explore the accuracy-FLOPs trade-off induced by the proposed method. 
Our results constantly stay above the Pareto frontier of previous works. 

\subsection{Importance of the orthogonal constraints}
\label{ssec:exporth}
Here we demonstrate the importance of adding the singular value orthogonality loss to the decomposed training process. We separately train two decomposed model with the same optimizer and hyperparameters, one with the orthogonality loss of $\lambda_o = 1.0$ and the other with $\lambda_o = 0$. No sparsity-inducing regularizer is applied to the singular values in this set of experiments. The experiments are conducted on ResNet-56 and ResNet-110 models, both trained under channel-wise decomposition and spatial-wise decomposition. The CIFAR-10 dataset is used for training and testing. As shown in Table~\ref{tab:orth}, the orthogonality loss enables the decomposed model to achieve similar or even better accuracy comparing to that of the original full model. On the contrary, training the decomposed model without the orthogonality loss will cause around 2\% accuracy loss. 

\begin{table}[tb]
    \caption{Comparison of top 1 accuracy on CIFAR-10 of ResNet models after full-rank SVD training, with or without orthogonality loss. [-Ch] means using channel-wise decomposition and [-Sp] means using spatial-wise decomposition. }
    \label{tab:orth}
    \centering
      \begin{tabular}{ccc}
      \toprule
      Model         &     $\lambda_o$     &      Accuracy (\%)\\
      \midrule
      ResNet-56     &     N/A             &      93.14      \\
      \midrule
      ResNet-56-Ch  &     1.0             &      93.28      \\
      ResNet-56-Ch  &     0.0             &      91.28      \\
      \midrule
      ResNet-56-Sp  &     1.0             &      93.36      \\
      ResNet-56-Sp  &     0.0             &      90.70      \\
      \bottomrule
      \toprule
      Model         &     $\lambda_o$     &      Accuracy (\%)\\
      \midrule
      ResNet-110     &     N/A             &      93.62     \\
      \midrule
      ResNet-110-Ch  &     1.0             &      93.58     \\
      ResNet-110-Ch  &     0.0             &      91.83     \\
      \midrule
      ResNet-110-Sp  &     1.0             &      93.93     \\
      ResNet-110-Sp  &     0.0             &      91.86     \\
      \bottomrule
      \end{tabular}
\end{table}

\subsection{Comparison of decomposition methods}
\label{ssec:expdec}

\begin{figure*}[tb]
    \centering
    \includegraphics[width=0.9\textwidth]{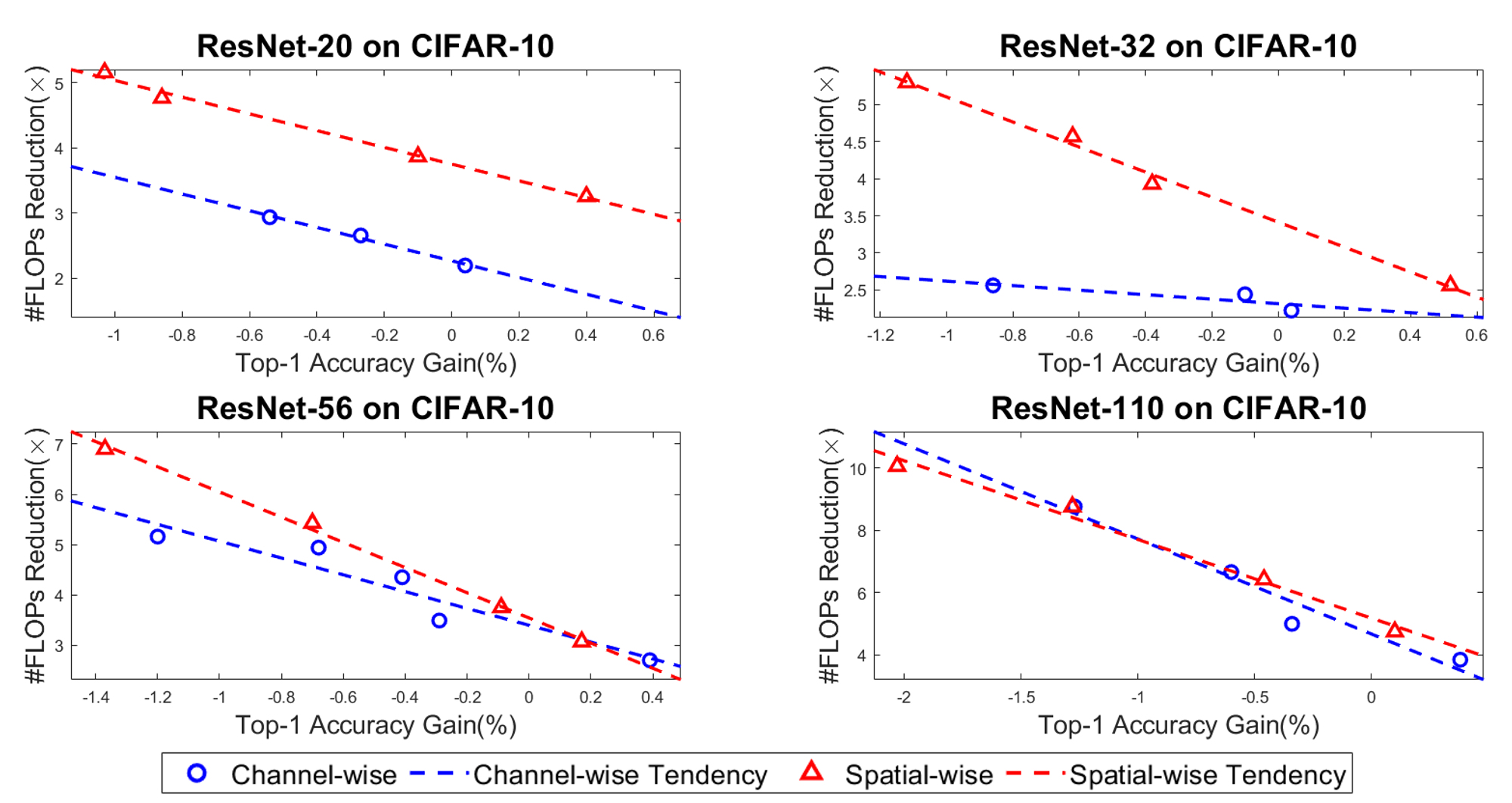}
    \caption{Effect of different decomposition methods. All models are achieved with Hoyer regularizer for singular value sparsity. Dash lines show the approximated tendency of the accuracy-compression tradeoff. See Appendix~\ref{ap:data} Table~\ref{tab:methods} for detailed data.}
    \vspace{-10pt}
    \label{fig:all_exp}
\end{figure*}

As mentioned in Section~\ref{ssec:dec}, we mainly consider the channel-wise and the spatial-wise decomposition method in this work. 
In this section, we compare the accuracy-\#FLOPs tradeoff tendency of the channel-wise decomposition and the spatial-wise decomposition. The tradeoff tendency of both decomposition methods are explored by training the decomposed model with Hoyer regularizer of different strength ($\lambda_s$ in Equation~(\ref{equ:obj})) on the singular values. The results are shown in Figure~\ref{fig:all_exp}. For shallower networks like ResNet-20 or ResNet-32 models, the spatial-wise decomposition shows a large advantage comparing to the channel-wise decomposition in the experiments, achieving significantly higher compression rate at similar accuracy. 
However, with a deeper network like ResNet-56 or ResNet-110, these two decomposition methods perform similarly. 
As discussed in Section~\ref{ssec:dec}, spatial-wise decomposition can utilize both spatial-wise redundancy and channel-wise redundancy, while the channel-wise decomposition utilizes channel-wise redundancy only. 
The observations in this set of experiments indicate that as DNN models get deeper, the channel-wise redundancy will become a dominant factor comparing to the spatial-wise redundancy. This corresponds to the fact that deeper layers in modern DNN typically have significantly more channels than shallower layers, resulting in significant channel-wise redundancy.

\subsection{Comparison of sparsity-inducing regularizers}
\label{ssec:expspar}

\begin{figure*}[tb]
    \centering
    \includegraphics[width=0.9\textwidth]{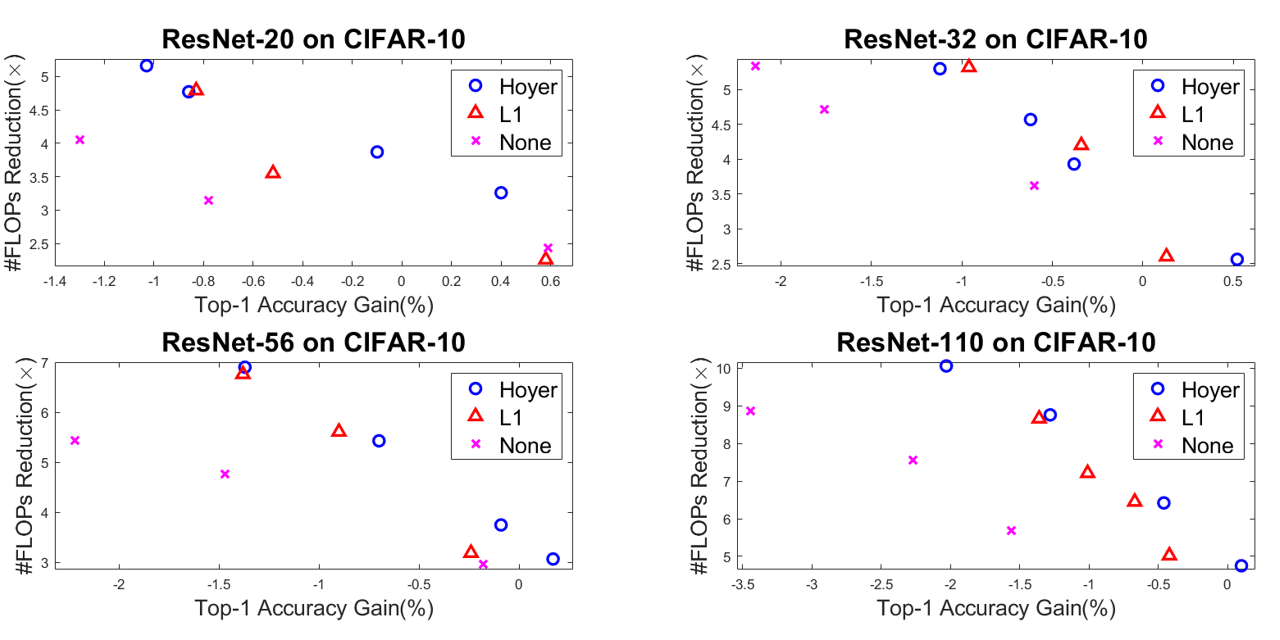}
    \caption{Effect of applying different sparsity-inducing regularizers during SVD training. All models are achieved with Spatial-wise decomposition. See Appendix~\ref{ap:data} Table~\ref{tab:methods} for detailed data.}
    \vspace{-10pt}
    \label{fig:reg}
\end{figure*}

Under the same model decomposition scheme, the main factor related to the final compression rate and the performance of the compressed model would be the choice of sparsity-inducing regularizers for the singular values. As mentioned in Section~\ref{ssec:spar}, we mainly consider the use of the $\normlone$ and the Hoyer regularizer in the proposed training scheme. 
In this section, we use spatial-wise decomposition setting to compare the effect of the $\normlone$ regularizer and the Hoyer regularizer. A controlled group is also trained with no sparsity-inducing regularizer applied during the SVD training.
The accuracy-\#FLOPs tradeoff is explored by changing the regularization strength and singular value pruning threshold. All other hyperparameters are kept the same during SVD training and fintuning process for all models.
Results are shown in Figure~\ref{fig:reg}.
The tradeoff tendency of the $\normlone$ regularizer constantly demonstrates a larger slope than that of the Hoyer regularizer. Under low accuracy loss, the Hoyer regularizer achieves a higher compression rate comparing to that of the $\normlone$ regularizer. However, if we are aiming for extremely high compression rate while allowing higher accuracy loss, the $\normlone$ regularizer can have a better performance. 
One possible reason for the difference in tendency is that the $\normlone$ regularizer will make all the singular values small through the training process, while the Hoyer regularizer will maintain the total energy of the singular values during the training, focusing more energy in larger singular values. Therefore more singular values can be removed from the decomposed model trained with the Hoyer regularizer without significantly hurting the performance of the model, resulting in higher compression rate at low accuracy loss. 
But it would be harder to keep most of the energy in a tiny amount of singular values than simply making everything closer to zero, therefore the $\normlone$ regularizer may perform better in the case of extremely high speedup. Comparing to the controlled group with no sparsity-inducing regularization, both the $\normlone$ regularizer and the Hoyer regularizer can achieve higher accuracy under similar compression rate, especially at high compression rate where the accuracy gap between with or without sparsity-inducing regularizer is more significant. Therefore applying sparsity-inducing regularizer on singular values is important for reaching a high performance low-rank model, as the weight will not naturally reach low-rank in training.

\subsection{Effectiveness of the overall training procedure}
\label{ssec:expoverall}
To show the ``full-rank SVD training, singular value pruning and low-rank finetuning'' training framework proposed in Section~\ref{ssec:train} is essential for reaching a high performance low-rank model, we take the model architecture of the low-rank models achieved from the proposed training procedure, reinitialize all the weights with random values, and train the low-rank model from scratch. For fair comparison, the reinitialized low-rank model is trained using the same training objective and hyperparameter choices as the low-rank finetuning step in our framework. As shown in Table~\ref{tab:spar}, with the same architecture and training process, training the low-rank model from scratch leads to around 2\% testing accuracy loss comparing to the accuracy achieved by the proposed training procedure. This result correspond to the fact that low-rank models are harder to train from scratch due to their low capacity~\cite{xu2018trained}. On the other hand, the full-rank SVD training step in our proposed framework provide sufficient capacity for the model to reach a high performance. Such high performance can still be preserved after singular value pruning, as the singular values are already sparse after the SVD training process.

\begin{table}[tb]
    \caption{Comparison of top 1 accuracy of the achieved low-rank ResNet models on CIFAR-10 vs. training the same model architecture from scratch. All low-rank models reported here are achieved with spatial-wise decomposition and Hoyer regularizer.}
    \label{tab:spar}
    \centering
      \begin{tabular}{ccc}
      \toprule
      Base Model         &     Training method     &      Accuracy (\%)\\
      \midrule
      ResNet-20                 &     Our method                &      91.39      \\
      Speed Up: 3.26$\times$    &     From scratch              &      89.43      \\
      \midrule
      ResNet-32                 &     Our method                &      91.76      \\
      Speed Up: 3.93$\times$    &     From scratch              &      90.55      \\
      \midrule
      ResNet-56                 &     Our method                &      93.27      \\
      Speed Up: 3.75$\times$    &     From scratch              &      91.55      \\
      \midrule
      ResNet-110                &     Our method                &      93.47      \\
      Speed Up: 6.42$\times$    &     From scratch              &      91.03      \\
      \bottomrule
      \end{tabular}
      \vspace{-10pt}
\end{table}

\subsection{Comparing with previous works}
\label{ssec:exppre}
\begin{figure*}[tb]
    \centering
    \includegraphics[width=0.85\textwidth]{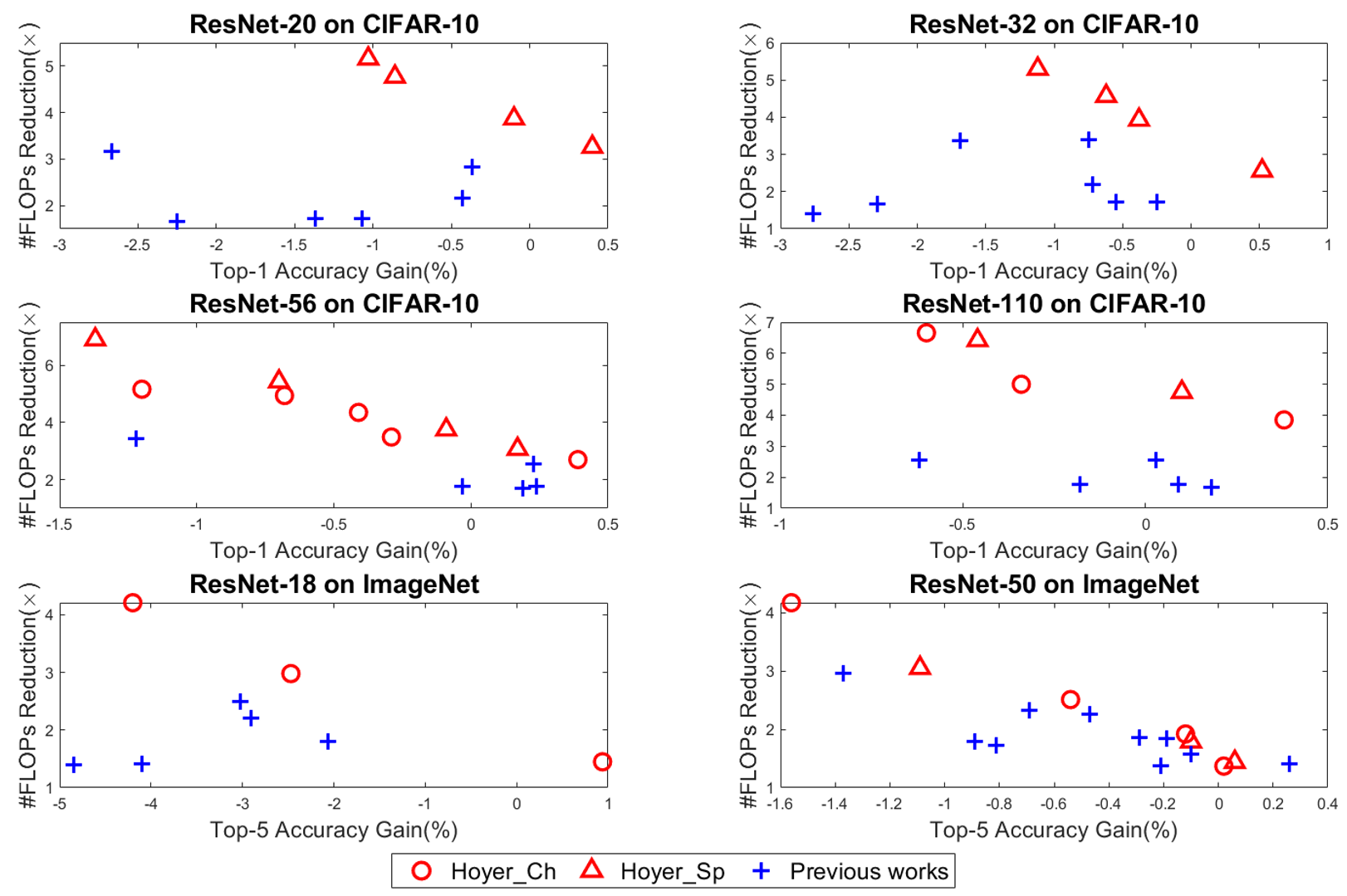}
    \caption{Comparison of accuracy-\#FLOPs tradeoff against previous methods. Being closer to the top-right corner indicates a better tradeoff point. See Appendix~\ref{ap:data} Table~\ref{tab:methods} -~\ref{tab:imagenet} for detailed results of our method, and Table~\ref{tab:comp} -~\ref{tab:comp_img} for all previous methods we compare to.}
    \label{fig:baseline}
    \vspace{-10pt}
\end{figure*}

We apply the proposed SVD training framework on the ResNet-20, ResNet-32, ResNet-56 and ResNet-110 models on the CIFAR-10 dataset as well as the ResNet-18 and ResNet-50 model on the ImageNet ILSVRC-2012 dataset to compare the accuracy-\#FLOPs tradeoff with previous methods. Here we mainly compare our method with state-of-the-art low-rank decomposition methods including Jaderberg et al.~\cite{jaderberg2014speeding}, Zhang et al.~\cite{zhang2015accelerating}, TRP~\cite{xu2018trained} and C-SGD~\cite{ding2019centripetal}, as well as recent filter pruning methods like NISP~\cite{yu2018nisp}, SFP~\cite{he2018soft} and CNN-FCF~\cite{li2019compressing}. The results of different models are shown in Figure~\ref{fig:baseline}. As analyzed in Section~\ref{ssec:expspar}, the spatial-wise decomposition methods achieves significantly higher compression rate than the channel-wise decomposition in shallower networks, while similar performance can be achieved when compressing a deeper model. Thus we compare the results of only the spatial-wise decomposition against previous works for ResNet-20 and ResNet-32. For other deeper networks, we report the results for both channel-wise and spatial-wise decomposition. As most of the previous works focus on compressing the model with a small accuracy loss, here we use the Hoyer regularizer for the singular values sparsity, as it can achieve a better compression rate than the $\normlone$ norm under low accuracy loss (see Section~\ref{ssec:expspar}). We use multiple strength for the Hoyer regularizer to explore the accuracy-\#FLOPs tradeoff, in order to compare against previous works with different accuracy levels. 
As shown in Figure~\ref{fig:baseline}, our proposed method can constantly achieve higher FLOPs reduction with less accuracy loss comparing to previous methods on different models and datasets. These comparison results prove that the proposed SVD training and singular value pruning scheme can effectively compress modern deep neural networks through low-rank decomposition.

\section{Conclusion}
\label{sec:con}

In this work, we propose the SVD training framework, which incorporates the full-rank decomposed training, singular value pruning and low-rank finetuning to reach low-rank DNNs with minor accuracy loss. We decompose each DNN layer to its full-rank SVD form before the training and directly train with the decomposed singular vectors and singular values, so we can keep an explicit measure of layers' ranks without performing the SVD on each step. 
Orthogonality regularizers are applied to the singular vectors during the training to keep the decomposed layers in a valid SVD form. And sparsity-inducing regularizers are applied to the singular values to explicitly induce low-rank layers. 

Thorough experiments are done to analyse each proposed technique. 
We demonstrate that the orthogonality regularization on singular vectors is crucial to the performance of the decomposed training process.
For decomposition methods, we find that the spatial-wise method performs better than channel-wise in shallower networks while the performances are similar for deeper models. 
For the sparsity-inducing regularizer, we show that higher compression rate can be achieved by Hoyer regularizer comparing to that of the $\normlone$ regularizer under low accuracy loss.
Our training framework is justified as training the low-rank model from scratch cannot reach the same accuracy achieved by our method.
We further apply the proposed method to various depth of ResNet models on both CIFAR-10 and ImageNet dataset, where we find our accuracy-\#FLOPs tradeoff constantly stays above the Pareto frontier of previous methods, including both factorization and structural pruning methods.
These results prove that this work provides an effective way for learning low-rank deep neural networks.

\subsubsection*{Acknowledgments}
This work was supported in part by NSF-1910299, NSF-1822085, DOE DE-SC0018064, and NSF IUCRC-1725456, as well as supports from Ergomotion, Inc.

{\small
\bibliographystyle{ieee_fullname}
\bibliography{iclr2020_conference}
}

\clearpage
\appendix
\section{Experiment setups}
\label{ap:setup}
Our experiments are done on the CIFAR-10 dataset~\cite{krizhevsky2009learning} and the ImageNet ILSVRC-2012 dataset~\cite{ILSVRC15}. We access both datasets via the API provided in the ``TorchVision'' Python package. As recommended in the PyTorch tutorial, we normalize the data and augment the data with random crop and random horizontal flip before the training. We use batch size 100 to train CIFAR-10 model and use 256 for the ImageNet model. For all the models on CIFAR-10, both the full-rank SVD training and the low-rank finetuning are trained for 164 epochs. The learning rate is set to 0.001 initially and decayed by 0.1 at epoch 81 and 122. For models on ImageNet, the full-rank SVD training is trained for 90 epochs, with initial learning rate 0.1 and learning rate decayed by 0.1 every 30 epochs. The low-rank finetuning is done for 60 epochs, starting at learning rate 0.01 and decay by 0.1 at epoch 30. We use pretrained full-rank decomposed model (trained with the orthogonality regularizer but without sparsity-inducing regularizer) to initialize the SVD training.
SGD optimizer with momentum 0.9 is used for optimizing all the models, with weight decay 5e-4 for CIFAR-10 models and 1e-4 for ImageNet models.
The accuracy reported in the experiment is the best validation accuracy achieved during the finetuning process. 

During the SVD training, the decay parameter of the orthogonality regularizer $\lambda_o$ is set to 1.0 for both channel-wise and spatial-wise decomposition on CIFAR-10. On ImageNet, for training the ResNet-18 model $\lambda_o$ is set to 5.0 for both decomposition methods. For the ResNet-50 model, $\lambda_o$ is set to 10.0 for channel-wise decomposition and 5.0 for spatial-wise decomposition. The decay parameter $\lambda_s$ for the sparsity-inducing regularizer and the energy threshold used for singular value pruning are altered through different set of experiments to fully explore the accuracy-\#FLOPs tradeoff. In most cases, the energy threshold is selected through a line search, where we find the highest percentage of energy that can be pruned without leading to a sudden accuracy drop. The $\lambda_s$ and the energy thresholds used in each set of the experiments are reported alongside the experiment results in Appendix~\ref{ap:data}.

\section{Detailed experiment results}
\label{ap:data}

In this section we list the exact data used to plot the experiment result figures in Section~\ref{sec:exp}. The results of our proposed method with various choice of decomposition method and sparsity-inducing regularizer tested on the CIFAR-10 dataset are listed in Table~\ref{tab:methods}. All of these data points are visualized in Figure~\ref{fig:all_exp} and Figure~\ref{fig:reg} to compare the tradeoff tendency under different conditions. As discussed in Section~\ref{ssec:exppre}, the results of spatial-wise decomposition with the Hoyer regularizer for ResNet-20 and ResNet-32 are shown in Figure~\ref{fig:baseline} to compare with previous methods. The results of both channel-wise and spatial-wise decomposition with the Hoyer regularizer are compared with previous methods in Figure~\ref{fig:baseline} for ResNet-56 and ResNet-110. For experiments on ImageNet dataset, the results of our method for the ResNet-18 model are listed in Table~\ref{tab:imagenet18}, and the results of our method for the ResNet-50 model are listed in Table~\ref{tab:imagenet}. 

The baseline results of previous works on compressing CIFAR-10 and ImageNet models used for comparison in Figure~\ref{fig:baseline} are listed in Table~\ref{tab:comp}-\ref{tab:comp_img}. As there are a large amount of previous works in this field, we only list the results of the most recent works here to show the state-of-the-art Pareto frontier. Therefore we choose state of the art low-rank compression methods like Jaderberg et al.~\cite{jaderberg2014speeding}, Zhang et al.~\cite{zhang2015accelerating}, TRP~\cite{xu2018trained} and C-SGD~\cite{ding2019centripetal}, as well as recent filter pruning methods like NISP~\cite{yu2018nisp}, SFP~\cite{he2018soft} and CNN-FCF~\cite{li2019compressing} as the baseline to compare our results against.

\clearpage
\onecolumn
\begin{longtable}{lccccc}
        \toprule
        Model       &Reg Type   &Decay&Energy Pruned&Accuracy Gain(\%)&Speed Up     \\
        \midrule
		 \endfirsthead
		 \endhead
		 \endfoot
        ResNet-20   & Hoyer     &0.03 &1.5e-5       &0.04             &2.20 $\times$\\
        Channel     &           &0.07 &6.0e-6       &-0.27            &2.66 $\times$\\
                    &           &0.1  &3.0e-6       &-0.54            &2.94 $\times$\\
                    \cline{2-6}
        Base Acc:   & L1        &0.01 &7.0e-2       &1.13             &1.43 $\times$\\
        90.93\%     &           &0.001&2.7e-2       &0.63             &1.59 $\times$\\
                    &           &0.1  &1.0e-1       &0.32             &2.10 $\times$\\
                    &           &0.3  &1.0e-1       &-0.48            &2.84 $\times$\\
                    \cline{2-6}
                    & None      &0.0  &1.9e-1       &-0.37            &2.03 $\times$\\
                    &           &0.0  &2.8e-1       &-0.52            &2.54 $\times$\\
                    &           &0.0  &3.3e-1       &-0.61            &2.88 $\times$\\
        \hline
        ResNet-20   & Hoyer     &0.01 &1.0e-3       &0.40             &3.26 $\times$\\
        Spatial     &           &0.03 &2.0e-5       &-0.10            &3.87 $\times$\\
                    &           &0.1  &4.0e-6       &-0.86            &4.77 $\times$\\
        Base Acc:   &           &0.01 &7.0e-3       &-1.03            &5.16 $\times$\\
                    \cline{2-6}
        90.99\%     & L1        &0.01 &6.0e-2       &0.58             &2.26 $\times$\\
                    &           &0.1  &1.0e-1       &-0.52            &3.55 $\times$\\
                    &           &0.3  &1.0e-1       &-0.83            &4.79 $\times$\\
                    \cline{2-6}
                    & None      &0.0  &2.9e-1       &0.59             &2.44 $\times$\\
                    &           &0.0  &3.9e-1       &-0.78            &3.15 $\times$\\
                    &           &0.0  &4.8e-1       &-1.30            &4.05 $\times$\\
        \hline
        ResNet-32   & Hoyer     &0.003&3.0e-3       &0.04             &2.22 $\times$\\
        Channel     &           &0.01 &1.0e-4       &-0.10            &2.44 $\times$\\
                    &           &0.03 &2.0e-6       &-0.86            &2.56 $\times$\\
                    \cline{2-6}
        Base Acc:   & L1        &0.03 &2.0e-2       &0.22             &1.58 $\times$\\
        92.12\%     &           &0.1  &1.0e-1       &-0.21            &2.84 $\times$\\
                    &           &0.3  &5.0e-2       &-0.96            &3.08 $\times$\\
                    \cline{2-6}
                    & None      &0.0  &1.8e-1       &-0.30            &2.23 $\times$\\
                    &           &0.0  &2.1e-1       &-0.11            &2.41 $\times$\\
                    &           &0.0  &2.3e-1       &-0.27            &2.51 $\times$\\
        \hline
        ResNet-32   & Hoyer     &0.001&5.0e-2       &0.52             &2.56 $\times$\\
        Spatial     &           &0.005&5.0e-3       &-0.38            &3.93 $\times$\\
                    &           &0.01 &8.0e-4       &-0.62            &4.57 $\times$\\
        Base Acc:   &           &0.03 &8.0e-6       &-1.12            &5.30 $\times$\\
                    \cline{2-6}
        92.14\%     & L1        &0.03 &7.0e-2       &0.13             &2.60 $\times$\\
                    &           &0.1  &2.5e-2       &-0.34            &4.20 $\times$\\
                    &           &0.1  &1.5e-1       &-0.96            &5.32 $\times$\\
                    \cline{2-6}
                    & None      &0.0  &3.8e-1       &-0.60            &3.62 $\times$\\
                    &           &0.0  &4.8e-1       &-1.76            &4.71 $\times$\\
                    &           &0.0  &5.3e-1       &-2.14            &5.34 $\times$\\
        \hline
        ResNet-56   & Hoyer     &0.001&2.0e-2       &0.39             &2.70 $\times$\\
        Channel     &           &0.003&1.0e-3       &-0.29            &3.49 $\times$\\
                    &           &0.01 &7.0e-6       &-0.41            &4.35 $\times$\\
        Base Acc:   &           &0.01 &2.0e-5       &-0.68            &4.94 $\times$\\
        93.28\%     &           &0.03 &3.0e-7       &-1.20            &5.16 $\times$\\
                    \cline{2-6}
                    & L1        &0.1  &3.0e-2       &-0.30            &4.25 $\times$\\
                    &           &0.1  &1.5e-1       &-0.59            &4.86 $\times$\\
                    \cline{2-6}
                    & None      &0.0  &2.8e-1       &-0.16            &2.91 $\times$\\
                    &           &0.0  &3.8e-1       &-0.98            &3.71 $\times$\\
                    &           &0.0  &4.7e-1       &-1.78            &4.70 $\times$\\
        \hline
        ResNet-56   & Hoyer     &0.001&3.0e-2       &0.17             &3.07 $\times$\\
        Spatial     &           &0.003&1.0e-3       &-0.09            &3.75 $\times$\\
                    &           &0.01 &1.0e-4       &-0.70            &5.43 $\times$\\
        Base Acc:   &           &0.03 &1.0e-6       &-1.37            &6.90 $\times$\\
                    \cline{2-6}
        93.36\%     & L1        &0.03 &5.0e-3       &-0.24            &3.19 $\times$\\
                    &           &0.03 &5.0e-2       &-0.90            &5.61 $\times$\\
                    &           &0.03 &2.5e-1       &-1.38            &6.76 $\times$\\
                    \cline{2-6}
                    & None      &0.0  &2.8e-1       &-0.18            &2.96 $\times$\\
                    &           &0.0  &4.7e-1       &-0.47            &4.76 $\times$\\
                    &           &0.0  &5.2e-1       &-2.22            &5.43 $\times$\\
        \hline
        ResNet-110  & Hoyer     &0.001&5.0e-3       &0.38             &3.85 $\times$\\
        Channel     &           &0.003&3.0e-4       &-0.34            &5.00 $\times$\\
                    &           &0.01 &3.0e-7       &-0.60            &6.66 $\times$\\
        Base Acc:   &           &0.03 &1.0e-6       &-1.27            &8.76 $\times$\\
                    \cline{2-6}
        93.58\%     & L1        &0.03 &1.0e-1       &-0.28            &5.02 $\times$\\
                    &           &0.03 &3.0e-1       &-1.27            &7.44 $\times$\\
                    \cline{2-6}
                    & None      &0.0  &3.7e-1       &-0.32            &4.26 $\times$\\
                    &           &0.0  &4.6e-1       &-1.86            &5.44 $\times$\\
                    &           &0.0  &5.5e-1       &-2.59            &7.03 $\times$\\
        \hline
        ResNet-110  & Hoyer     &0.001&1.3e-2       &0.10             &4.75 $\times$\\
        Spatial     &           &0.003&7.0e-4       &-0.46            &6.42 $\times$\\
                    &           &0.01 &2.0e-5       &-1.28            &8.76 $\times$\\
        Base Acc:   &           &0.03 &2.0e-8       &-2.03            &10.06$\times$\\
                    \cline{2-6}
        93.93\%     & L1        &0.03 &3.0e-2       &-0.42            &5.02 $\times$\\
                    &           &0.03 &1.0e-1       &-0.67            &6.45 $\times$\\
                    &           &0.03 &1.5e-1       &-1.01            &7.21 $\times$\\
                    &           &0.03 &2.5e-1       &-1.36            &8.66 $\times$\\
                    \cline{2-6}
                    & None      &0.0  &4.7e-1       &-1.56            &5.69 $\times$\\
                    &           &0.0  &5.6e-1       &-2.27            &7.55 $\times$\\
                    &           &0.0  &6.1e-1       &-3.44            &8.87 $\times$\\
        \bottomrule
        
\caption{Full results of applying the proposed method on ResNet models on the CIFAR-10 dataset with various hyperparameters. [Decay] marks the decay variable for the sparse regularization, i.e. $\lambda_s$. [Energy Pruned] means the energy threshold used for singular value pruning, i.e. $e$. [Accuracy Gain] denotes the gain of Top-1 accuracy from the accuracy of the baseline full-rank model. [Speed Up] is computed as the ratio of \#FLOPs of the original model and the achieved low-rank model.}
\label{tab:methods}
\end{longtable}

\begin{table}[htbp]
    \centering
    \caption{Results of applying the proposed method on ResNet-18 model on the ImageNet dataset. Hoyer regularizer is used as the sparsity-inducing regularizer for the singular values. Top-5 validation accuracy is reported in the [Base Acc] and the [Accuracy Gain] columns. [Speed Up] is computed as the ratio of \#FLOPs of the original model and the achieved low-rank model.}
    \label{tab:imagenet18}
    \begin{tabular}{lccccc}
         \toprule
        Decompose   & Base Acc  &Decay&Energy Pruned&Accuracy Gain&Speed Up     \\
        \midrule
        Channel     & 88.54\%   &0.002 &5.0e-4       &0.94\%             &1.45 $\times$\\
                    &           &0.003 &1.0e-4       &-1.28\%            &2.03 $\times$\\
                    &           &0.005 &1.0e-4       &-2.47\%            &2.98 $\times$\\
                    &           &0.01  &1.0e-5       &-4.20\%            &4.21 $\times$\\
        \hline
        Spatial     & 88.54\%   &0.002 &1.0e-4       &0.67\%             &1.61 $\times$\\
                    &           &0.005 &1.0e-4       &-0.84\%            &2.98 $\times$\\
                    &           &0.01  &1.0e-4       &-3.13\%            &6.36 $\times$\\
        \bottomrule
    \end{tabular}
    
\end{table}

\begin{table}[htbp]
    \centering
    \caption{Results of applying the proposed method on ResNet-50 model on the ImageNet dataset. Hoyer regularizer is used as the sparsity-inducing regularizer for the singular values. Top-5 validation accuracy is reported in the [Base Acc] and the [Accuracy Gain] columns. [Speed Up] is computed as the ratio of \#FLOPs of the original model and the achieved low-rank model.}
    \label{tab:imagenet}
    \begin{tabular}{lccccc}
         \toprule
        Decompose   & Base Acc  &Decay&Energy Pruned&Accuracy Gain&Speed Up     \\
        \midrule
        Channel     & 91.72\%   &0.001 &1.0e-4       &0.02\%             &1.37 $\times$\\
                    &           &0.002 &1.0e-4       &-0.12\%            &1.92 $\times$\\
                    &           &0.003 &5.0e-5       &-0.54\%            &2.51 $\times$\\
                    &           &0.005 &5.0e-5       &-1.56\%            &4.17 $\times$\\
        \hline
        Spatial     & 91.91\%   &0.0005&1.0e-3       &0.06\%             &1.44 $\times$\\
                    &           &0.001 &1.0e-4       &-0.10\%            &1.79 $\times$\\
                    &           &0.002 &2.0e-4       &-1.09\%            &3.05 $\times$\\
        \bottomrule
    \end{tabular}
    
\end{table}

\begin{table}[htbp]
    \centering
    \caption{Baselines on the CIFAR-10 dataset. [Accu.$\uparrow$] means the Top-1 accuracy gain comparing to that of the full model. [Sp. Up] denotes speed up computed as the ratio of \#FLOPs before and after the model compression. [-] is marked when no result is available in the paper.}
     \label{tab:comp}
        \begin{tabular}{l|cc|cc|cc|cc}
        \toprule
        \multirow{2}*{Method}&\multicolumn{2}{c|}{ResNet-20}&\multicolumn{2}{c|}{ResNet-32}&\multicolumn{2}{c|}{ResNet-56}&\multicolumn{2}{c}{ResNet-110}\\
        \cline{2-9}
        &Accu.$\uparrow$&Sp. Up&Accu.$\uparrow$&Sp. Up&Accu.$\uparrow$&Sp. Up&Accu.$\uparrow$&Sp. Up\\
        \hline
        Zhang et al. &-3.61\% &1.41 $\times$&-2.76\%&1.41 $\times$&   -   &     -      &   -   &     -      \\
        \hline
        Jaderberg et al. &-2.25\% &1.66 $\times$&-2.29\%&1.68 $\times$&   -   &     -      &   -   &     -      \\
        \hline
        TRP-Ch &-0.43\% &2.17 $\times$&-0.72\%&2.20 $\times$&   -   &     -      &   -   &     -      \\
        \hline
        TRP-Sp &-0.37\% &2.84 $\times$&-0.75\%&3.40 $\times$&   -   &     -      &   -   &     -      \\
        \hline
        SFP    &-1.37\% &1.79 $\times$&-0.55\%&1.71 $\times$&0.19\% &1.70 $\times $&0.18\% &1.69 $\times$\\
        \hline
        \multirow{2}*{CNN-FCF}&-1.07\% &1.71 $\times$&-0.25\%&1.73 $\times$&0.24\% &1.75 $\times$&0.09\% &1.76 $\times$\\
        \cline{2-9}
               &-2.67\% &3.17 $\times$&-1.69\%&3.36 $\times$&-1.22\%&3.44 $\times$&-0.62\%&2.55 $\times$\\
        \hline
        C-SGD-5/8&   -  &     -      &   -   &     -      &0.23\% &2.55 $\times$&0.03\% &2.56 $\times$\\
        \hline
        Nisp   &   -    &     -      &   -   &     -      &-0.03\%&1.77 $\times$&-0.18\%&1.78 $\times$\\
        \bottomrule
        \end{tabular}
\end{table}

\begin{table}[htbp]
    \caption{Baselines of compressing ResNet-18 model on the ImageNet dataset. [Accu.$\uparrow$] means the Top-5 accuracy gain comparing to that of the full model. [Sp. Up] denotes speed up computed as the ratio of \#FLOPs before and after the model compression. }
    \label{tab:comp_img18}
    \centering
    \begin{tabular}{ccc|ccc}
    \toprule
    \multicolumn{3}{c|}{Channel-wise} & \multicolumn{3}{c}{Spatial-wise}\\
    \midrule
    Method  & Accu.$\uparrow$  & Sp. Up   & Method  & Accu.$\uparrow$  & Sp. Up  \\
    \midrule
    Zhang et al.& -4.85\%   & $1.39 \times$ & Jaderberg et al. & -4.82\%   & $2.00 \times$  \\
    Zhang et al.& -4.10\%   & $1.41 \times$ & TRP-Sp      & -1.80\%   & $2.60 \times$  \\
    TRP-Ch      & -2.06\%   & $1.81 \times$ & TRP-Sp      & -2.71\%   & $3.20 \times$  \\
    TRP-Ch      & -2.91\%   & $2.20 \times$ & TRP-Sp      & -3.24\%   & $3.68 \times$  \\
    TRP-Ch      & -3.02\%   & $2.50 \times$ &    &     &    \\
    \bottomrule
    \end{tabular}
\end{table}

\begin{table}[htbp]
    \caption{Baselines of compressing ResNet-50 model on the ImageNet dataset. [Accu.$\uparrow$] means the Top-5 accuracy gain comparing to that of the full model. [Sp. Up] denotes speed up computed as the ratio of \#FLOPs before and after the model compression. }
    \label{tab:comp_img}
    \centering
    \begin{tabular}{ccc|ccc}
    \toprule
    Method  & Accu.$\uparrow$  & Sp. Up   & Method  & Accu.$\uparrow$  & Sp. Up  \\
    \midrule
    SFP         & -0.81\%   & $1.72 \times$ & NISP-50-A & -0.21\%   & $1.38 \times$  \\
    CNN-FCF-A   & +0.26\%   & $1.41 \times$ & NISP-50-B & -0.89\%   & $1.79 \times$  \\
    CNN-FCF-B   & -0.19\%   & $1.85 \times$ & C-SGD-70  & -0.10\%   & $1.58 \times$  \\
    CNN-FCF-C   & -0.69\%   & $2.33 \times$ & C-SGD-50  & -0.29\%   & $1.86 \times$  \\
    CNN-FCF-D   & -1.37\%   & $2.96 \times$ & C-SGD-30  & -0.47\%   & $2.26 \times$  \\
    \bottomrule
    \end{tabular}
\end{table}

\clearpage
\twocolumn

\end{document}